# MobiFace: A Novel Dataset for Mobile Face Tracking in the Wild

Yiming Lin, Shiyang Cheng, Jie Shen and Maja Pantic
Intelligent Behaviour Understanding Group, Department of Computing
Imperial College London
{yiming.lin15, shiyang.cheng11, jie.shen07}@imperial.ac.uk
maja.pantic@gmail.com

***Abstract*— Face tracking serves as the crucial initial step in mobile applications trying to analyse target faces over time in mobile settings. However, this problem has received little attention, mainly due to the scarcity of dedicated face tracking benchmarks. In this work, we introduce MobiFace, the first dataset for single face tracking in mobile situations. It consists of 80 unedited live-streaming mobile videos captured by 70 different smartphone users in fully unconstrained environments. Over $95K$ bounding boxes are manually labelled. The videos are carefully selected to cover typical smartphone usage. The videos are also annotated with 14 attributes, including 6 newly proposed attributes and 8 commonly seen in object tracking. 36 state-of-the-art trackers, including facial landmark trackers, generic object trackers and trackers that we have fine-tuned or improved, are evaluated. The results suggest that mobile face tracking cannot be solved through existing approaches. In addition, we show that fine-tuning on the MobiFace training data significantly boosts the performance of deep learning-based trackers, suggesting that MobiFace captures the unique characteristics of mobile face tracking. Our goal is to offer the community a diverse dataset to enable the design and evaluation of mobile face trackers. The dataset, annotations and the evaluation server will be on `https://mobiface.github.io/`.**

## I. INTRODUCTION

Face analysis on mobile platforms has attracted increasing levels of interest in recent years [5]–[8]. One crucial element is face tracking, which, given its initial location specified in the first frame, finds the bounding box of the target face in a video. Despite recent attempts to address this problem [9]–[11], tracking a face on mobile platforms remains extremely challenging due to the large appearance variations caused by illumination changes, out-of-plane rotations, heavy occlusions, target disappearance from the camera view among various difficulties.

One important prerequisite of tackling this challenge is the collection of a large scale in-the-wild mobile dataset for developing and evaluating face trackers. However, existing mobile datasets [1], [2], [6] fail to serve this purpose, as they are (i) not designed for face tracking, and; (ii) not collected in fully unconstrained environments. To our knowledge the only in-the-wild mobile dataset is the one proposed by Yu and Ramamoorthi [4]. Unfortunately, this dataset is quite small, consisting of only $8,756$ frames, and does not provide bounding box annotations. In this work, we attempt to bridge this gap by proposing the first mobile face tracking dataset in the wild. Our dataset contains 80 unedited mobile live-streamed videos with $95,635$ frames with bounding box annotations provided for all the frames.

Although it might appear that the mobile face tracking problem can be readily solved using existing generic object tracking methods, extensive experiments on our dataset show this to be incorrect. Four key differences between the two problems illustrate why. First, the target faces can undergo large scale variations due to the mobility of smartphones, whereas the target's size and the aspect ratio rarely change in most object tracking videos. Second, due to the use of hand-held devices in the mobile footage, the motion of the target can be fast and sometimes unpredictable. Third, rarely does object tracking have similar objects in the same video, whereas in mobile face tracking the tracker can often encounter multiple faces. Finally, due to the smaller field of view of mobile cameras, targets can be easily occluded or out of view. Nevertheless, domain adaptation from generic object tracking to face tracking can still provide a promising starting point given a sufficient amount of data [12]–[14].

Following the trend of deep learning in computer vision, the research community has employed data-driven approaches to solve a wide range of face-related problems, such as face detection [15] and facial landmark tracking [16], [17]. In this paper, we introduce MobiFace, the first large scale mobile dataset dedicated for face tracking, as a significant step towards solving the mobile face tracking problem.

Our contribution can be summarised as:

1) We introduce MobiFace which consists of 80 unedited in-the-wild mobile video uploaded by 70 smartphone users. We provide bounding box annotations for all the $95,635$ frames. We also define 14 attributes for these mobile videos and provide annotations for each one.
2) We benchmark 36 state-of-the-art tracking methods and models, including 4 facial landmark trackers, 14 object trackers and 18 trackers that we improved or fine-tuned. The results indicate that mobile face tracking remains challenging for all existing methods.
3) We demonstrate that fine-tuning on MobiFace significantly boosts the performance of deep learning-based trackers. This suggests MobiFace captures the unique characteristics of mobile face tracking, demonstrating its use potential for the research community.

| Datasets | Capture device | # of videos | # of frames | # of labelled frames | Avg. frames | Edited | Application | Year |
|---|---|---|---|---|---|---|---|---|
| MOBIO [1] | Nokia N93i & MacBook | 28,800 | 5,490,000 | 0 | 190 | No | Biometrics | 2008 |
| AA-01-FD [2] | iPhone 5s | 750 | 8,036 | 8,036 | 11 | Yes | Detection | 2015 |
| UMDAA-02-FD [3] | Nexus 5 | 8,756 | 33,209 | 33,209 | 4 | Yes | Detection | 2016 |
| Yu and Ramamoorthi [4] | Various smartphones | 33 | 8,318 | 0 | 252 | No | Video stabilisation | 2018 |
| **MobiFace (this paper)** | **Various smartphones** | **80** | **95,635** | **95,635** | **1,195** | **No** | **Tracking** | **2018** |

TABLE I: A summary of recent publicly available face datasets in the mobile domain. The column 'Edited' denotes whether or not the source videos are downsampled or composed of multiple shots.

4) All videos, training data annotations and the evaluation server are provided on our project website. To guarantee a fair comparison, we reserve MobiFace test set annotations on our evaluation server. Researchers are encouraged to submit their tracking results for benchmarking.

## II. Related Work

### A. Face Datasets in the Mobile Domain

Despite the tremendous efforts in building large scale face databases [17]–[20], there has been a scarcity of datasets tailored for mobile face tracking in the wild. The MOBIO [21] dataset is one of the earliest attempts to collect mobile videos for face and speaker verification. In this work, 160 participants were recruited and recorded in 12 different sessions. However their videos were collected in the controlled lab environment, where the subjects were required to position their heads inside a fixed region during the recording. Similar mobile datasets for face verification also exist, with more variations on background and less restrictions on head pose. The AA-01-FD [2] dataset and UMDAA-02-FD [3] dataset are two widely used mobile face datasets. Nevertheless, their videos are often very short and contain only one large main face, which makes them less useful for developing mobile face trackers or benchmarking object trackers in the mobile domain. Yu and Ramamoorthi [4] recently proposed a mobile dataset that contains 33 in-the-wild selfie videos. At initial look the data set would seem suitable for benchmarking trackers in mobile settings. However, the size of dataset is too small, and no face bounding box annotation is provided.

Through MobiFace, our work aims to fill this gap. To our knowledge, this is the first mobile dataset dedicated to face tracking in the wild. This dataset exhibits rich variations in camera quality, pose, illumination and background. Table I compares our MobiFace dataset with other existing mobile datasets.

### B. Face Tracking

In contrast with facial landmark tracking [17], a face tracker provides only the target face's bounding box, which can then be used as the input for facial landmark localisation and other high level face analysis tasks. This work focuses on addressing face tracking in mobile situations. Specifically, given the initial state of a target face in the first frame, a face tracker should continuously estimate the states, *e.g.*, presence, size and location, of the target face in all subsequent frames.

Face-TLD [9] was one of the earliest attempts to apply the tracking-by-detection diagram to face tracking. [10] employed an incremental principal component analysis (PCA) model to obtain robust face representation. Due to the lack of face tracking benchmarks the performance of these trackers was reported either on a few videos [22] or on small subsets from object tracking benchmarks [11], which yielded results unable to be directly compared. Our proposed dataset aims to provide the community with an unified benchmark for the development of mobile face trackers.

### C. Single Object Tracking

Single object tracking, or object tracking, shares a similar objective with face tracking. In contrast to the slow advancements in face tracking, object tracking has progressed rapidly, due to the availability of many visual tracking databases and competitions, including VOT [13], OTB [12] and VisDrone [14], and others [23]–[25]. As face is an instance of the object class, object tracking methods can potentially be adapted to mobile face tracking, provided that sufficient training videos are made available.

*1) Correlation filter trackers:* Correlation Filter (CF) [26] has been widely used to develop accurate and fast trackers. This is primarily due to modelling of target's movement as a circulant matrix, which represents the dense sampling of the target. The circulant shift formulation results in very fast solution to the underlying ridge regression problem in the Fourier domain. The KCF tracker [27] proposed to map the input features to a high dimensional space by the kernalised CF. DSST [28] introduced a multi-scale search space to handle large object scale variations. Staple [29] combined colour features and histogram of oriented gradients (HOG) features to enhance feature robustness. Most recently, the HCF [30] tracker exploited different features from multiple layers of a pre-trained deep convolutional networks in the CF framework, achieving promising results.

*2) Deep trackers:* Following the success of Convolutional Neural Networks (CNNs) in computer vision applications, a large number of object trackers that employ deep models were proposed. We have broadly categorised these trackers according to their tracking strategies. Readers can refer to [31] for a thorough review of deep trackers.

**Tracking-by-detection**. These trackers train an online classifier to discriminate the target from the background using the samples collected during tracking. MDNet [32] applied a multi-domain offline training technique to train the full network, and fine-tuned the last fully connected layer online. CREST [33] reformulated CF as a fully connected layer for tracking and used online residual learning to train the CF layer. VITAL [34] explored adversarial learning to augment the feature extraction process, obtaining better features to capture large appearance changes. MetaTracker [35] introduced an offline meta-learning method to adjust the initial network for online adaptation.

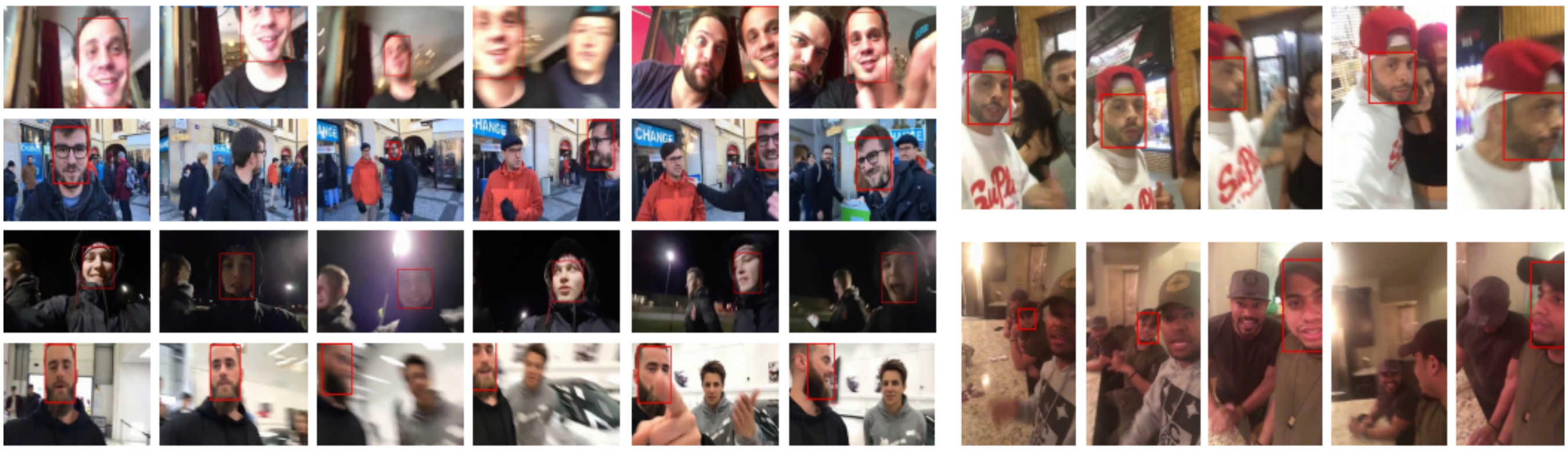

Fig. 1: Exemplar videos from our proposed MobiFace. Red rectangles indicate the ground truth bounding boxes.

**Template matching**. The template matching-based tracker devises two identical CNNs to extract features from the target image patch and the search region respectively, which are combined to generate a response map with maximum value indicating the target position. The pioneering SiamFC [36] adopted a pre-trained fully convolutional Siamese network for feature extraction. It was trained end-to-end and operated beyond real-time. CFNet [37] exploited CF in training, which allowed shallow CNNs to extract robust features for tracking. SiamRPN [38] introduced a region proposal module to propose candidate regions of different sizes and aspect ratios, able to deal with dramatic scale changes of the target object. MemTrack [39] proposed a Long Short-Term Memory (LSTM) module to control the template adaptation to capture the appearance changes along the time.

## III. MobiFace Dataset

In this section, we introduce MobiFace, the very first mobile dataset tailored for face tracking in the wild. MobiFace contains 80 videos curated from thousands of mobile live-streaming videos uploaded by different internet users. Fig. 1 shows some examples of the videos. We manually annotate all the $95,635$ frames with a bounding box around the target face. We also annotate each mobile video with 14 attributes, of which, 6 are newly proposed attributes commonly seen in mobile situations. We provide details about our data collection and annotation procedure as well as the evaluation metrics. We also highlight the new challenges posed by this novel dataset in this section.

### A. Data Collection

We used YouTube Data API to fetch videos from YouTube mobile live-streaming channel, which is maintained daily by YouTube. Those videos are usually raw footage recorded and uploaded by different vloggers worldwide, without any video editing or visual effects. Most of them are captured under fully unconstrained environments, such as a protest event, a wild party or even a wedding, where interactions of the user with the environment/audience are natural. Exemple videos that show the user interacting with the surrounding environment can be found in the $2^{nd}$ and $4^{th}$ sequence on the left of Fig. 1 while typical videos exhibiting the user-audience interaction can be found in the $1^{st}$ and $3^{rd}$ sequence.

To create a large pool of candidate videos, we subscribed and followed the aforementioned channel for three consecutive months. In total, we have collected 6,201 unedited videos for later selection. Among these videos, we went through each one of them and discarded those that do not capture a face on the camera, *e.g.*, live-streaming of a street. We further refined our list by two criteria: (i) the target face should at least appear in 10% of the video frames; (ii) to serve the purpose of visual tracking, the target face should not always stay still. As a result, we picked 80 video sequences, with each video carefully segmented to retain the main part that best reflects the challenging scenarios in the mobile domain.

As facial images are of critical privacy, we concern ourselves with the legal issue on circulating the data within academia. Thus we have contacted the video owners for their consents to use the videos for academic purposes.

### B. Annotation for Tracking

We provide a manually annotated face bounding box for all the frames in MobiFace. For each sequence, we annotate the initial bounding box and define strict annotation protocols for the annotators to label the successive frames. We recruited three experienced annotators and developed a cross-platform web application for the annotation process. Our annotation protocol is that, two annotators are assigned different videos for the bounding box annotation, while the third annotator serves as the quality control person, who is committed to review the annotation and report the possible confusions to us. In this manner, we can quickly resolve the annotation problem and give feedback to the annotators.

We also derived clear rules to guarantee the quality and consistency of our annotation process. Different from object tracking that may not require a tight bounding box, we tried to provide an accurate close-fitting box that can benefit the latter face analysis. Henceforth, four rules were specified as follows:

| Attribute | # of videos | Description |
|---|---|---|
| FC (Frontal Camera) | 61 | The video was captured using the frontal camera. |
| RC (Rear Camera) | 18 | The video was captured using the rear camera. |
| PT (Portrait) | 9 | The orientation of the video is portrait. |
| LS (Landscape) | 71 | The orientation of the video is landscape. |
| CS (Camera Switch) | 1 | The camera switches from frontal to rear in the video or vice verse. |
| MF (Multiple Faces) | 64 | More than one face exist in the view. |
| IV (Illumination Variation) | 38 | Significant illumination change on the target face. |
| SV (Scale Variation) | 31 | The area ratio of two bounding boxes in two consecutive frames is smaller than 0.7. |
| OCC (Occlusion) | 50 | The face is partially or fully occluded. |
| FM (Fast Motion) | 19 | The displacement of the target centre is larger than the frame width between consecutive frames. |
| IPR (In-Plane Rotation) | 31 | The face rotates in the image plane. |
| OPR (Out-of-Plane Rotation) | 77 | The face rotates out of the image plane. |
| OV (Out-of-View) | 72 | Part or all of the target leaves the view. |
| BL (Blur) | 52 | The target face is blurred due to the motion of target or smartphone, out-of-focus and low resolution. |

TABLE II: MobiFace video attributes. Top: six proposed attributes (see Sec. III-C). Bottom: eight common attributes as in [12].

*1) Bounding box:* An upright bounding box is labelled tightly around the target face in each frame. We required it to cover the forehead and the chin but not the ears. For profile faces, the bounding box should include the nose tip and exclude the ear. We do not use eclipse annotations [40] since most face-related algorithms, *e.g.*, facial landmarks tracking, take an upright bounding box as input.

*2) Occlusion and out-of-view:* Some benchmarks annotate only the visible part of the target even if it is severely occluded [41]–[43]. We argue that this introduces drastic scale changes between frames and may cause bias in the evaluation results. Considering the fact that a small portion of the face does not contribute to the successive face analysis, we mark the state that 90% of the target face is occluded or out-of-view as *absent*, and require the tracker to report all zero values to represent the absent state.

*3) Rotation:* The faces with over 120 degrees of out-of-plane rotation are also considered *absent* (see $2^{nd}$ sequence of Fig. 1). For in-plane rotations, we require the bounding boxes to be upright and cover the whole face.

*4) Large faces:* When the faces are extremely close to the camera and display only some visible parts, *e.g.*, eyes and nose, we annotate the whole visible part as it is difficult to estimate the extent of the bounding box outside the view.

### C. Attributes Annotation

Apart from the bounding box annotation, each video is also annotated with a list of high-level attributes defined in Table II. We introduce *six novel and unique attributes* that are commonly seen in the mobile scenarios but not considered in object tracking benchmarks.

*1) Frotnal Camera (FC), Rear Camera (RC) and Camera Switch (CS):* These three attributes relate to the camera of mobile phone. **FC** and **RC** indicate whether the video is captured by the frontal or the rear camera. The frontal camera is more frequently used in selfie videos while the rear camera is usually for general-purpose shooting. Because of this fact, the recorded videos captured from the frontal and the rear camera pose different challenges for tracking. For example, the frontal camera often has a closer view of the phone holder, which may not only result in a bigger face, but also an incomplete/occluded face. On the other hand, the rear camera usually captures a smaller face with many background noises or distractors. **CS** suggests the case that the user switches the camera from the frontal camera to the rear ones, or vice versa. The challenges of this attribute include: (i) two cameras often have different specs such as megapixels, aperture and focal length; (ii) the switch of camera is sudden and unpredictable. Although we do not provide many videos with this extremely difficult attribute, it is certainly of our future interest to collect more such data.

*2) Portrait (PT) and Landscape (LS):* They denote the orientation of the video. Depending on the user's holding position, the captured video can be either in *portrait* or *landscape* orientation.

*3) Multiple Faces (MF):* It indicates whether there are more than one face in the video or not. There are usually no similar objects in one single video in object tracking datasets [12]–[14], [25] and mobile datasets [1]–[4].

The remaining eight attributes are commonly used in object tracking and we annotate them following the protocols defined in [12]. Full distribution of attributions can be found on the project website.

### D. Evaluation Metrics

We utilise three conventional metrics, *i.e.*, precision plot, success plot and frame per second (FPS), to quantitatively evaluate a face tracker on the proposed MobiFace dataset.

*1) Normalised precision plot:* Precision plot [12] is a widely used evaluation metric on tracking. Precision is described by the euclidean distance between the centre location of the tracked face and the ground truth box. However, as our unedited mobile videos differ greatly in resolution, we adopt the recently proposed normalised precision value [43] in our benchmark. The size of the frame is used for the normalisation, and we rank the trackers based on the area under the curve (AUC) for normalised precision value between 0 and 0.5.

*2) Success plot:* The success plot [12] shows the percentage of frames in which the intersection of union (IoU) of the predicted and ground truth bounding box is greater than a given threshold. Denoting the ground truth bounding box as $r_{gt}$ and the predicted bounding box as $r_p$, the IoU metric is defined as $\frac{r_{gt} \cap r_p}{r_{gt} \cup r_p}$, where $\cap$ and $\cup$ represent the intersection and union of two regions, respectively. The threshold value ranges from 0 to 1. A representative score for each tracker is the AUC of the success plot.

*3) FPS:* This is the average speed of the evaluated tracker running across all the sequences. The initialisation time is not considered here, thus we count the FPS from the beginning of frame-to-frame tracking. Ideally, a mobile face tracker should run at a high FPS (on CPU or GPU) to allow potential migration to the actual mobile devices. Due to a lack of available face/object tracking implementations on mobile devices, we can only provide the FPS on desktop environment which should still be indicative to the efficiency of the trackers.

## IV. Benchmark Results

In this section, we evaluate the effectiveness and efficiency of many different trackers on our novel MobiFace dataset, and provide in-depth discussion on the current issue in mobile face tracking. All of our experiments were conducted on a desktop machine with a Intel i9-7900X CPU (3.30GHz) and one GTX 1080 Ti GPU.

### A. Evaluated Trackers

We gathered and evaluated 18 representative tracking methods [9], [27], [29], [30], [32]–[39], [44]–[47] as listed in Table III on our dataset. Among these trackers, 12 of them are *short-term* trackers from the object tracking benchmarks [12], [13], 2 trackers are *long-term* trackers, while the remaining 4 trackers are state-of-the-art 2D and 3D facial landmark trackers [46], [47]. Unfortunately, there are not many trackers designed for long-term tracking, we hence equipped all 4 facial landmark trackers [46], [47] and 3 short-term trackers [32], [35], [36] with a simple re-detection module to alleviate this problem. In order to demonstrate the usefulness of MobiFace, we fine-tuned these trackers on the training set, and compared them with the versions fine-tuned using YouTube Faces [18] on the test set. Note that we also apply the re-detection mechanism to those fine-tuned trackers to get the best possible result on mobile face tracking. In total, we have benchmarked 36 different tracking methods and models.

*1) Short-term trackers:* Table III summarises the evaluated short-term trackers. Since these trackers were designed for *generic objects*, we expected that their performance could generalise to some extend to faces.

*2) Long-term trackers:* Long-term trackers are normally equipped with a re-detection module which is called when they believe the target is lost because of the severe occlusion or tracking drift. We first consider **TLD** [9] and **EBT** [49] as they were designed for long-term tracking. Unfortunately, EBT does not output the absent state and we do not have access to the source code for evaluation, only TLD is available for us. Besides, we assembled a simple tracker, dubbed **DVNet**, by concatenating a robust face detector [48] and a state-of-the-art face verification method [45]. Although DVNet is not a real tracker as it ignores temporal information, we use it here to illustrate that mobile face tracking cannot be fully solved by this naive combination and thus more systematic designs are required.

| Trackers | Feature | Method | Device | Implementation | Long-term |
|---|---|---|---|---|---|
| KCF [27] | HOG | **CF** | CPU | **C++** | No |
| Staple [29] | HOG+Color | **CF** | CPU | **M** | No |
| ECO [44] | CNN | **CF** | GPU | **MCN** | No |
| HCF [30] | CNN | **CF** | GPU | **MCN** | No |
| SiamFC [36] | CNN | **TM** | GPU | **MCN** | No |
| CFNet [37] | CNN | **TM** | GPU | **MCN** | No |
| VITAL [34] | CNN | **TD** | GPU | **P+P** | No |
| CREST [33] | CNN | **TD** | GPU | **P+P** | No |
| MemTrack [39] | CNN | **TM** | GPU | **P+P** | No |
| MDNet [32] | CNN | **TD** | GPU | **P+P** | No |
| MetaMDNet [35] | CNN | **TD** | GPU | **P+P** | No |
| DaSiamRPN [38] | CNN | **TM** | GPU | **P+P** | No |
| TLD [9] | HOG | **TD** | CPU | **M** | Yes |
| $CMHM_{2D}$ [46] | HOG | **TD** | GPU | **P+T** | No |
| $CMHM_{3D}$ [46] | HOG | **TD** | GPU | **P+T** | No |
| $FAN_{2D}$ [47] | HOG | **TD** | GPU | **P+P** | No |
| $FAN_{3D}$ [47] | HOG | **TD** | GPU | **P+P** | No |
| DVNet [45], [48] | CNN | **TD** | GPU | **P+T** | Yes |

TABLE III: Evaluated state-of-the-art trackers. The initials stand for: **CF** - Correlation Filter, **TD** - Tracking-by-Detection, **TM** - Template Matching, **M** - Matlab, **MCN** - MatConvNet, **P+P** - Python and PyTorch, **P+T** - Python and Tensorflow.

Furthermore, we augment MDNet [32], SiamFC [36] and MetaMDNet [35] with a re-detection logic similar to that described in [41]. Specifically, if the maximum score of the response is below a threshold, the tracker should enter the *absent* state. During this state, it will search around the last position in the subsequent frames until the maximum score surpasses the threshold. The threshold is updated in a similar manner as that of the conventional template update scheme in the tracking literature [36], [37]. We denote these modified trackers as **MDNet+R**, **SiamFC+R** and **MetaMDNet+R**.

*3) Facial landmark trackers:* Although MobiFace is not aimed at benchmarking non-rigid face tracking methods, two 2D landmark trackers **$CMHM_{2D}$** [46] and **$FAN_{2D}$** [47] and their 3D variants[1], **$CMHM_{3D}$** and **$FAN_{3D}$**, are included in the evaluation for completeness. These trackers have produced state-of-the-art results on recent 2D and 3D face alignment benchmarks [16], [47]. For each sequence, we initialised the tracker with the ground truth bounding box in the first frame, then used the current fitting result as the initialisation of the next frame. For tracking evaluation, the trackers used the bounding box predicted in the previous frame. Additionally, we implemented a similar re-detection scheme as that of long-term trackers to the landmark trackers. We used the sum of response scores from the landmark trackers as the confidence score, for which a threshold is empirically set to decide if a face detector [48] should be called to re-initialise the tracker. The score is also updated during tracking using the aforementioned template update scheme. We denoted these modified trackers as **$CMHM_{2D}$+R**, **$CMHM_{3D}$+R**, **$FAN_{2D}$+R** and **$FAN_{3D}$+R**.

### B. Benchmark Results on MobiFace

*1) Short-term trackers:* Fig. 2 shows the results of short-term trackers on the whole MobiFace dataset. The top performing tracker is ECO [44] (47.0% in the success plot), which uses a factorised convolution operator to reduce parameters for online training. Nevertheless, none of the tested trackers achieved over 50% in success rate, whereas in the other object tracking benchmark [12], they can easily score over 80%. The same situation can be observed from the normalised precision plot, compared with the generally high values (around 70%) reported in OTB100 [12], [43], the normalised precision scores reported in our dataset are still quite low. These facts suggest that there is still room for improvement in the mobile face tracking problem.

[1]The facial landmarks returned by these 3D trackers are in fact the 2D projections of the 3D facial landmarks.

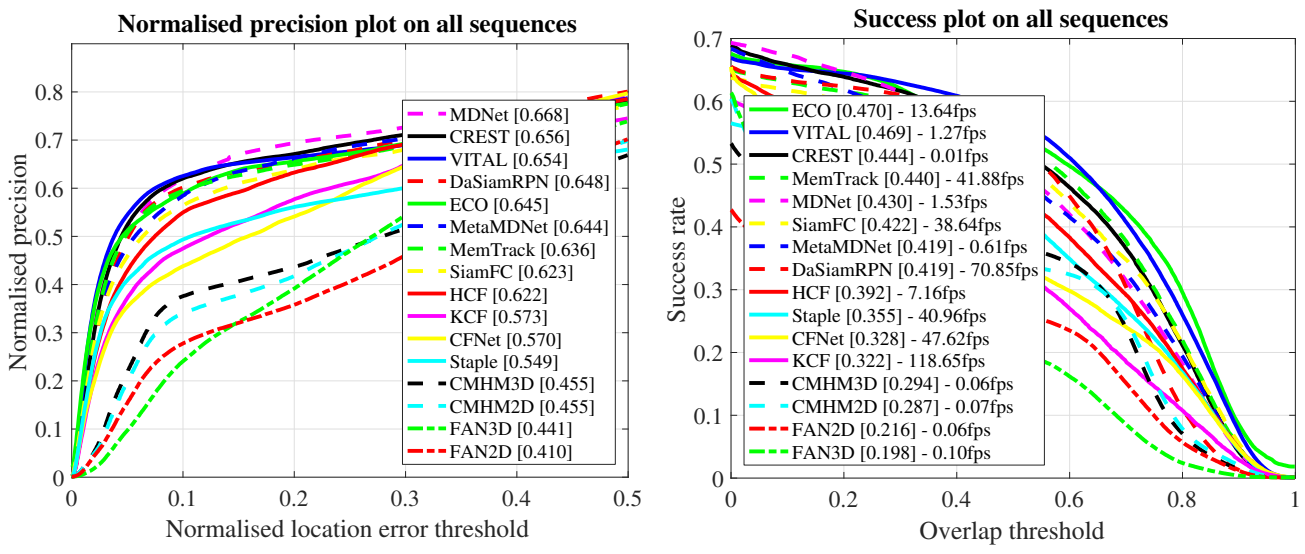

Fig. 2: Results of short-term trackers on MobiFace.

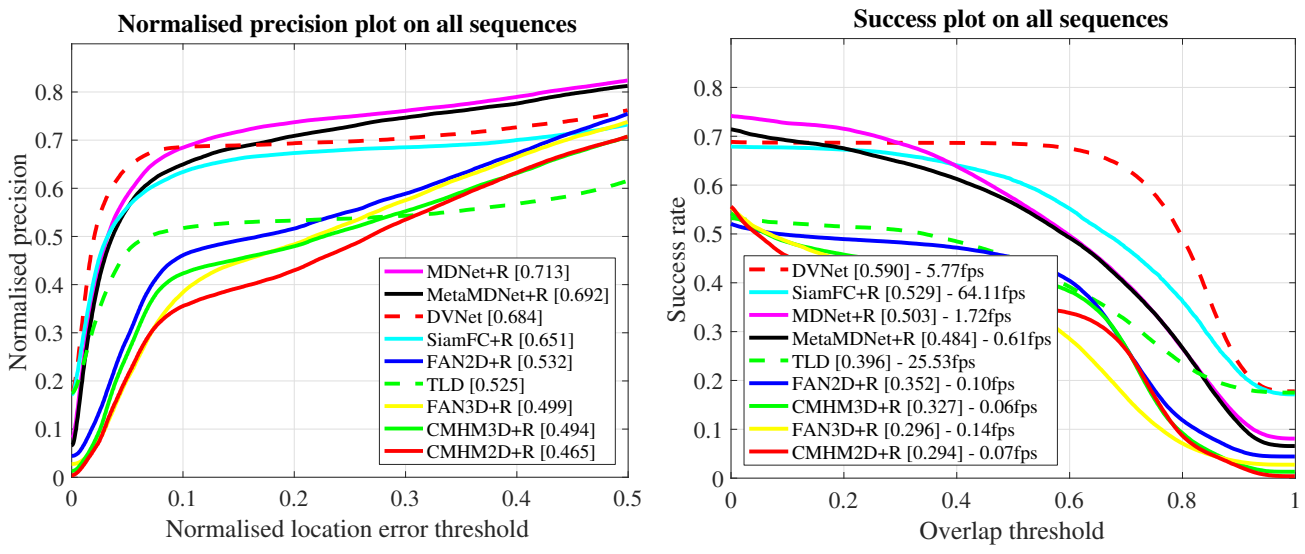

Fig. 3: Results of long-term trackers on MobiFace.

It is worth noting that although the template-matching trackers SiamFC, DaSiamFC and MemTrack produce slightly worse results (42.2%, 41.9%, and 44.0% respectively), they are significantly faster than the other trackers. Nevertheless, one common drawback of the short-term trackers is that they cannot identify the absent state, thus may fail to re-locate the out-of-view target when it reenters the scene. In the next experiment, we will show that with a simple additional re-detection module, the performance of short-term trackers can be improved by a considerable margin.

*2) Long-term trackers:* Fig. 3 shows the results of long-term trackers on all MobiFace videos. Although DVNet is seemingly the best method, it is by definition not a tracking algorithm, and the run-time speed is quite slow. Even worse, its success rate stalls at around 0.7 regardless of any lower overlap thresholds. This suggests that the naive combination of face detection and verification is not an effective and elegant solution to the mobile face tracking problem.

Notice that SiamFC+R benefits from a re-detection module in terms of accuracy and speed, as the re-detection helps SiamFC avoid incorrect and unnecessary update of the template. SiamFC+R runs at one magnitude faster than DVNet with comparable results. In the light of this fact, we believe that a more systematic design of template-matching strategies and re-detection modules can be a promising future direction.

*3) Facial landmark trackers:* Fig. 2 and 3 both show that landmark trackers CMHM and FAN cannot compete with most of the objects tracker, with or without re-detection module. This is because landmark trackers are very error-prone to motion blur and occlusion, while object trackers are more robust on these cases.

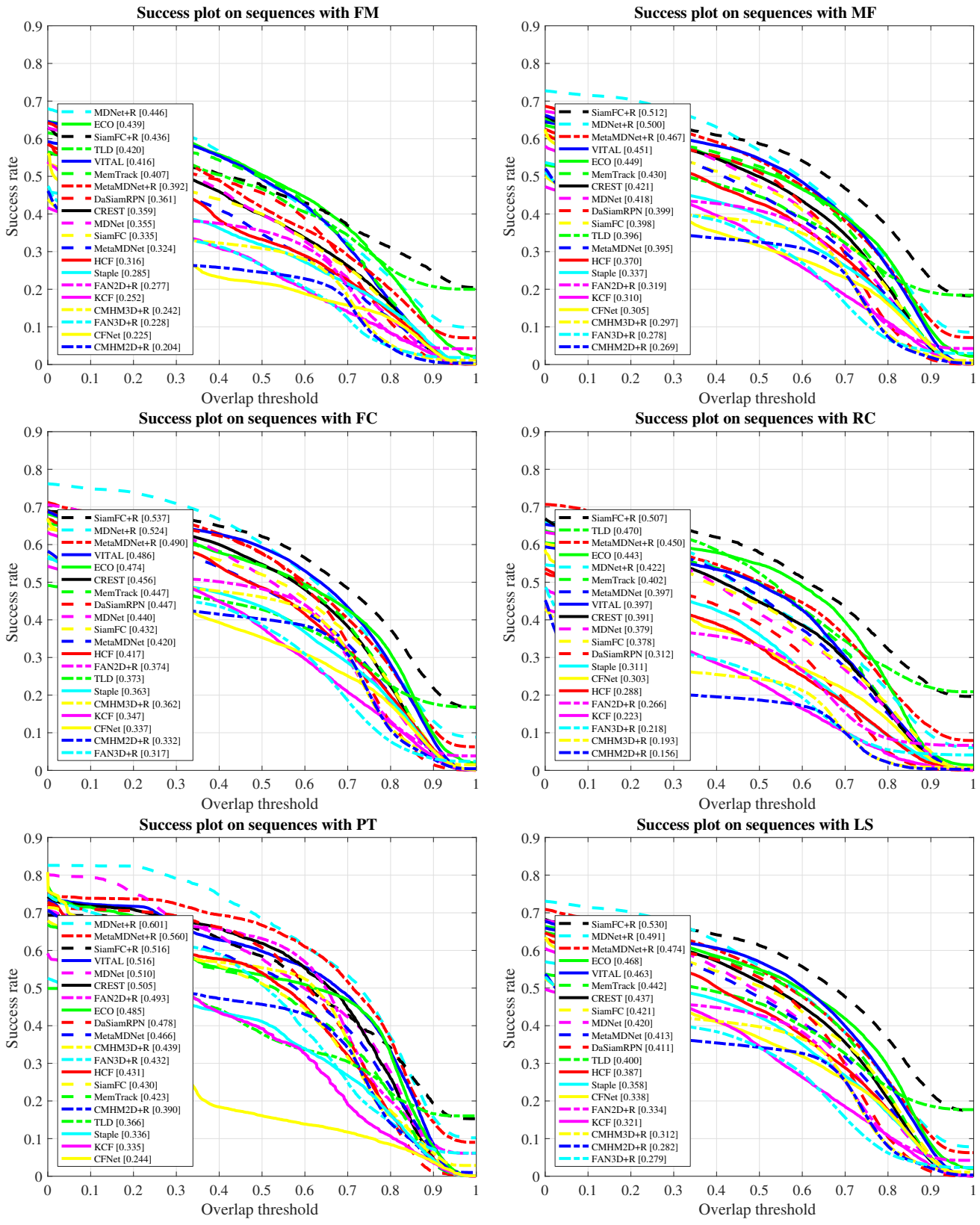

Fig. 4: Results of trackers on six different attributes, *i.e.*, *Fast Motion*, *Multiple Faces*, *Frontal Camera*, *Rear Camera*, *Portrait* and *Landscape*.

## C. Attribute Specific Results

Each video in MobiFace is labelled with some of the 14 attributes as described in Sec. III-C. According to the attributes of a video, we categorise the results from the short-term and long-term trackers and plot them in separate figures. Due to the page limit, we only highlight five new attributes (*i.e.*, Frontal Camera (FC), Rear Camera (RC), Landscape (LS), Portrait (PT) and Multiple Faces (MF)), plus the Fast Motion (FM) that is prevalent in mobile footage. We refer the readers to our project website for the complete results.

Overall, *Fast Motion* and *Rear Camera* cause the most troubles for the tested trackers. Taking SiamFC+R as the reference method, its success rate for the dataset is 52.9% (Fig. 3), while for the FM and RC sequences (Fig. 4), it drops to 43.6% and 50.7% respectively. Similar observations can be made in the other methods such as MDNet+R and ECO. The reason of RC being troublesome to the trackers is that the rear camera usually gives a distant view of the object, thus any minor movement of the phone can result in a fast magnified movement of the object in the video.

All these facts imply that fast motion is one of the biggest challenges in mobile face tracking. We further analyse how the unique attributes proposed in MobiFace may affect trackers in Sec. IV-D.2.

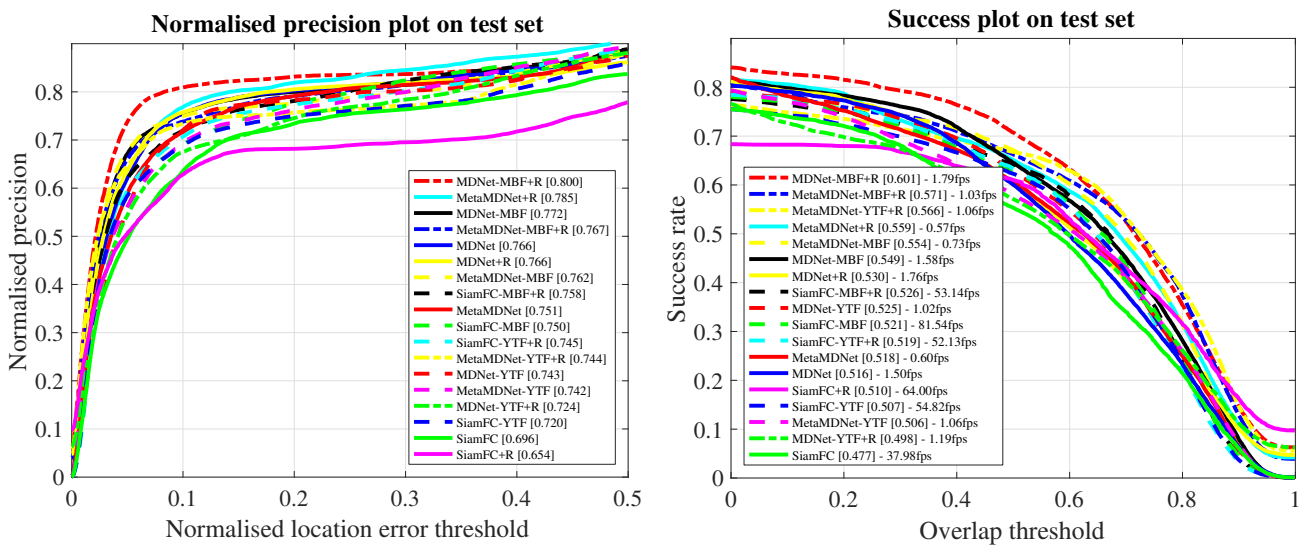

Fig. 5: Fine-tuning on MobiFace improves trackers.

### D. Retraining on MobiFace

*1) Fine-tuning deep trackers:* Deep neural networks normally require fine-tuning to adapt to other domains. Most deep trackers we evaluated were developed for generic objects and should be fine-tuned for the faces. We randomly split the whole MobiFace dataset into the train and the test set with a ratio of 8:2. This partition is *subject independent*, as we ensure there is no overlap of subjects between two sets. To provide fair comparison, the test set annotations will be held privately and can only be accessed by our evaluation server. We fine-tuned MDNet, SiamFC, MetaMDNet, MDNet+R, SiamFC+R and MetaMDNet+R on the train set. The resulting trackers are named by adding a suffix **-MBF**. To show that MobiFace captures the characteristics of mobile face tracking, we also fine-tuned those trackers on YouTube Faces [18] and reported their performances. These fine-tuned trackers are named with a suffix **-YTF**.

Fig. 5 shows the results of fine-tuned trackers on the MobiFace test set. It can be seen that fine-tuning with MobiFace constantly improves the trackers over their original versions, for example, MDNet-MBF, gains an improvement of 0.033 over MDNet in the success rate. On the contrary, fine-tuning on YouTube Faces dataset [18] delivers an inferior and inconsistent performance. In the success plot, a prominent example is that, MetaMDNet-YTF+R overshadows MetaMDNet+R by 0.7%, while MetaMDNet-YTF (50.6%) performs worse than MetaMDNet (51.8%).

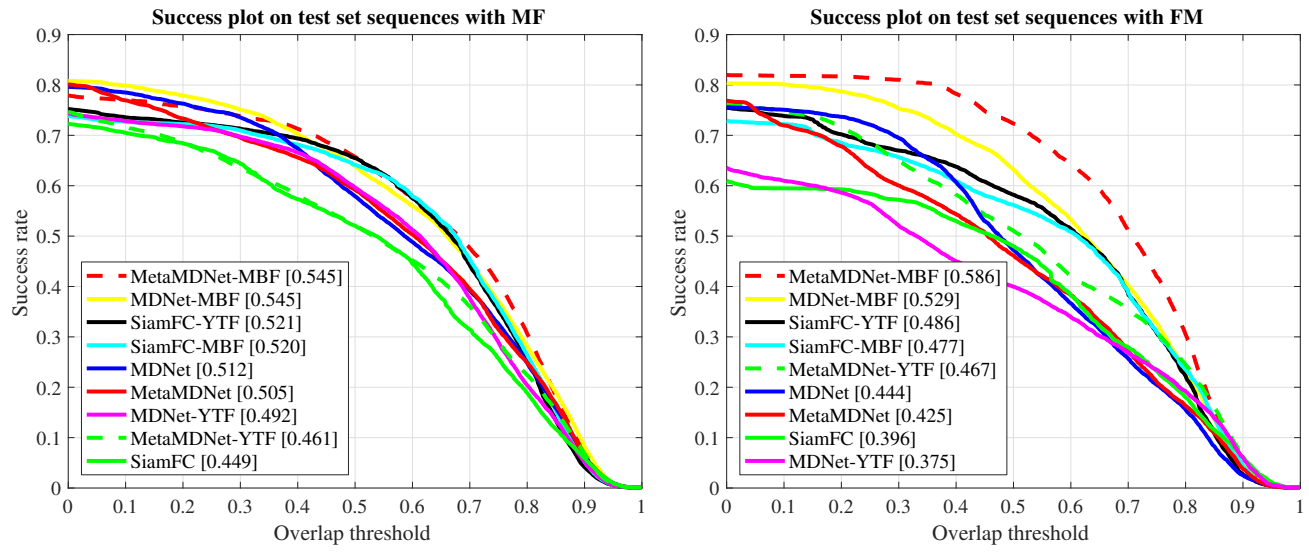

Fig. 6: Fine-tuning on MobiFace improves trackers in Multiple Faces and Fast Motion scenarios. See Sec.IV-D.2.

*2) How fine-tuning helps:* To gain more insights about the performance improvement obtained from fine-tuning on MobiFace, we show the attribute-wise performance in Fig. 6. Interestingly, fine-tuning on MobiFace train set provides the most significant improvement on sequences with *Multiple Faces* and *Fast Motion*. This is likely owing to the fact that these two attributes seldom appear in the other datasets except our MobiFace. On the other hand, fine-tuning on YouTube Faces [18] actually hurts the performance on these two attributes. The possible reason is that videos in YouTube Faces usually contain a single face and the shooting device is almost always still. These two attributes deserve more attention, as they are challenging and prevalent in the mobile footage. Arguably, MobiFace lays a good foundation for resolving these issues in mobile face tracking.

*3) Qualitative results:* Fig. 7 show qualitative results of the variants of SiamFC [36] on the MobiFace test set. It is worth emphasising that after fine-tuning on MobiFace, the tracker can better handle profile faces, and the scenarios of multiple faces and fast motion. These qualitative results comply with our observations from the previous experiments.

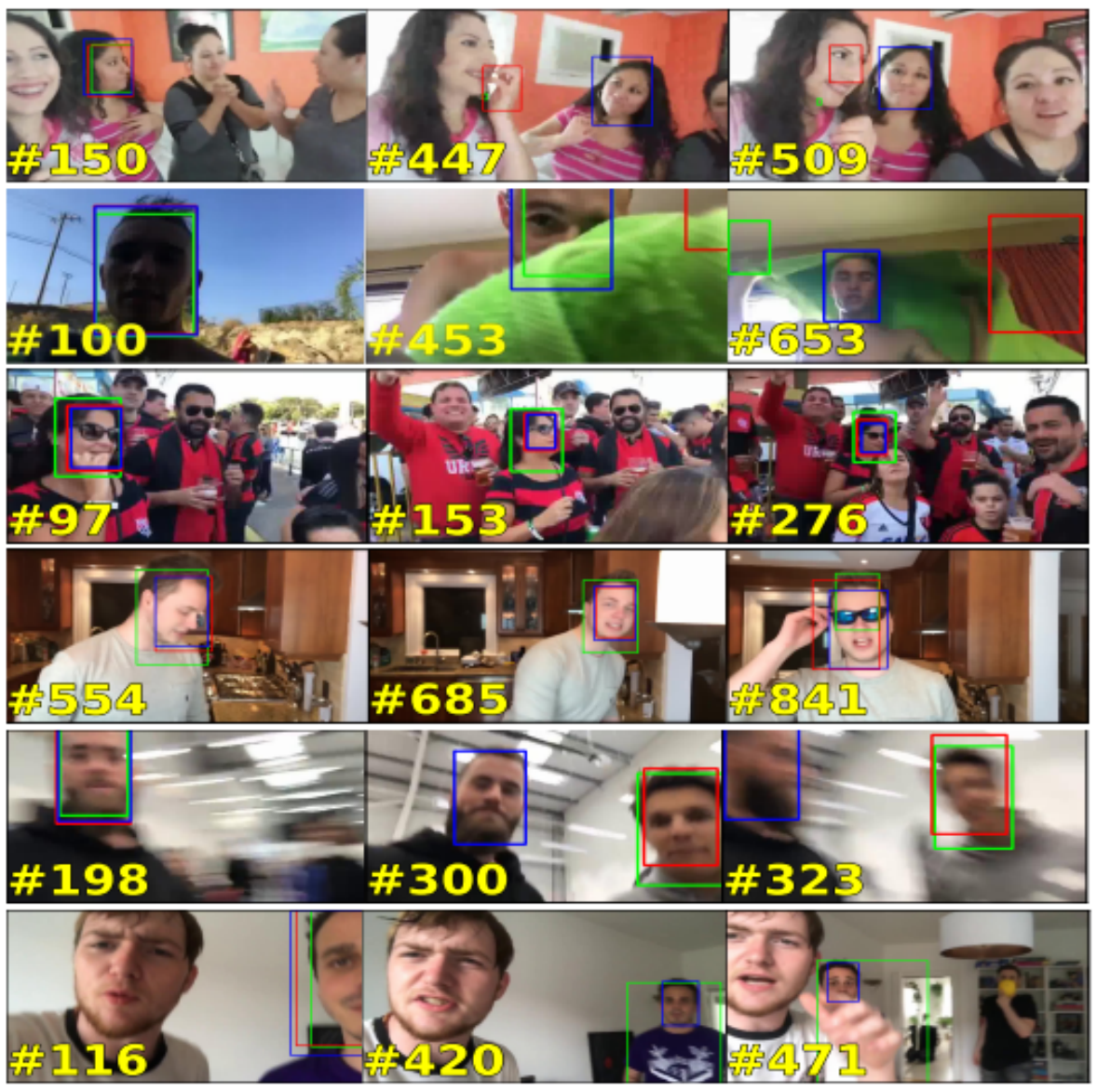

Fig. 7: Qualitative results comparing variants of SiamFC. SiamFC-MBF was fine-tuned on MobiFace training set and SiamFC-YTF fine-tuned on YouTube Faces (Sec. IV-D.3).

## V. Conclusion

We present MobiFace, the first mobile dataset tailored for developing and benchmarking single face trackers. This dataset consists of 80 *in-the-wild* mobile videos from 70 smartphone users, with a total of 95,635 frames. We provide a manually annotated bounding box for each frame, and define 14 video attributes for the mobile videos, of which, 6 are newly proposed.

We have conducted a comprehensive evaluation of 36 advanced tracking methods and models on MobiFace. Our results suggest that mobile face tracking is still a very challenging problem that cannot be fully solved by existing landmark or object trackers, neither by a simple concatenation of face detection and verification method. Despite this, we take a step towards solving this problem by showing that fine-tuning on MobiFace substantially boosts the performance of the trackers. This experiment not only demonstrates the usefulness of MobiFace, but also highlights the richness of its contents in the mobile domain.

All the MobiFace videos, the training set annotations and the evaluation library will be made publicly available on https://mobiface.github.io/. To ensure a fair comparison, we reserve the test set annotations on the server, where researchers can upload their results for evaluation.